\documentclass{article}

\usepackage[preprint, nonatbib]{nips_2018}

\usepackage{graphicx}

\usepackage[utf8]{inputenc} 
\usepackage[T1]{fontenc}    
\usepackage{hyperref}       
\usepackage{url}            
\usepackage{booktabs}       
\usepackage{amsfonts}       
\usepackage{nicefrac}       
\usepackage{microtype}      

\usepackage{float}
\usepackage{cleveref}
\usepackage[group-separator={\ }]{siunitx}
\usepackage{algorithm,algorithmicx,algpseudocode}
\usepackage{textcomp}

\newcommand{\argmax}{\mathop{\rm argmax}}

\title{Convolutional CRFs for Semantic Segmentation}

\author{
  Marvin T. T. Teichmann\thanks{Primary Contact} \\
  Department of Engineering\\
  University of Cambridge \\
  Cambridge, United Kingdom\\
  \texttt{mttt2@eng.cam.ac.uk} \\
  \And
  Roberto Cipolla \\
  Department of Engineering\\
  University of Cambridge \\
  Cambridge, United Kingdom\\
  \texttt{cipolla@eng.cam.ac.uk} \\
  \And
}

\begin{document}

\maketitle


\begin{abstract}

For the challenging semantic image segmentation task the most efficient models have traditionally combined the structured modelling capabilities of Conditional Random Fields (CRFs) with the feature extraction power of CNNs. In more recent works however, CRF post-processing has fallen out of favour. We argue that this is mainly due to the slow training and inference speeds of CRFs, as well as the difficulty of learning the internal CRF parameters. To overcome both issues we propose to add the assumption of conditional independence to the framework of fully-connected CRFs. This allows us to reformulate the inference in terms of convolutions, which can be implemented highly efficiently on GPUs. Doing so speeds up inference and training by two orders of magnitude. All parameters of the convolutional CRFs can easily be optimized using backpropagation. To facilitating further CRF research we make our implementation publicly available. Please visit: \url{https://github.com/MarvinTeichmann/ConvCRF}
\end{abstract}

\section{Introduction}




Semantic image segmentation, which aims to produce a categorical label for each pixel in an image, is a very import task for visual perception. Convolutional Neuronal Networks have been proven to be very strong in tackling semantic segmentation tasks \cite{long2015fully,chen2018deeplab,chen2017rethinking,zhao2017pyramid}. While deep neural networks are extremely powerful in extracting local features and performing good predictions utilizing a small field of view, they lack the capability to utilize global context information and cannot model interactions between predictions directly. Thus it has been suggested that simple feed-forward CNNs my not be the perfect model for structured predictions tasks such as semantic segmentation \cite{zhao2017pyramid,lin2016efficient,zheng2015conditional}. Several authors have successfully combined the effectiveness of CNNs to extract powerful features, with the modelling power of CRFs in order to address the discussed issues \cite{lin2016efficient, chandra2016fast, zheng2015conditional}. Despite their indisputable success, structured models have fallen out of favour in more recent approaches \cite{DBLP:journals/corr/WuSH16e,chen2017rethinking,zhao2017pyramid}.

We believe that the main reasons for this development are that CRFs are notoriously slow and hard to optimize. Learning the features for the structured component of the CRF is an open research problem \cite{vemulapalli2016gaussian,lin2016efficient} and many approaches rely on entirely hand-crafted Gaussian features \cite{DBLP:journals/corr/abs-1210-5644,zheng2015conditional,DBLP:journals/corr/SchwingU15,chen2018deeplab}. In addition, CRF inference is two orders of magnitude slower then CNN inference. This makes CRF based approaches to slow for many practical applications. The long training times of the current generation of CRFs also make more in-depth research and experiments with such structured models impractical.

To solve both of these issue we propose to add the strong and valid assumption of conditional independence to the existing framework of fully-connected CRFs (FullCRFs) \cite{DBLP:journals/corr/abs-1210-5644}. This allows us to reformulate a large proportion of the inference as convolutions, which can be implemented highly efficiently on GPUs. We call our method convolutional CRFs (ConvCRFs). Backpropagation \cite{rumelhart1986learning} can be used to train all parameters of the ConvCRF. Inference in ConvCRFs can be performed in less then \num{10}{ms}. This is a speed increase of two-orders of magnitude compared to FullCRFs. We believe that those fast train and inference speeds will greatly benefit future research and hope that our results help to revive CRFs as a popular method to solve structured tasks.

\section{Related Work}

Recent advances in semantic segmentation are mainly driven by powerful deep neural network architectures \cite{krizhevsky2012imagenet,simonyan2014very,he2016deep,DBLP:journals/corr/WuSH16e}. Following the ideas of FCNs \cite{long2015fully}, transposed convolution layers are applied at the end of the prediction pipeline to produce high-resolution output. Atrous (dilated) convolutions \cite{DBLP:journals/corr/ChenPKMY14,yu2015multi} are commonly applied to preserve spatial information in feature space.

Many architectures have been proposed \cite{noh2015learning, ronneberger2015u,badrinarayanan2017segnet,paszke2016enet,teichmann2016multinet,DBLP:journals/corr/WuSH16e}, based on the ideas above. All of those approaches have in common that they primarily rely on the powerful feature extraction provided by CNNs. Predictions are pixel-wise and conditionally independent (given the common feature base of nearby pixels). Structured knowledge and background context is ignored in these models.

One popular way to integrate structured predictions into CNN pipelines is to apply a fully-connected CRF (FullCRF) \cite{DBLP:journals/corr/abs-1210-5644} on top of the CNN prediction \cite{DBLP:journals/corr/ChenPKMY14,zheng2015conditional,DBLP:journals/corr/SchwingU15,lin2016efficient,chandra2016fast}. Pyramid pooling \cite{zhao2017pyramid} is a proposed alternative to CRFs, incorporating some context knowledge. Pyramid pooling increases the effective field of view of the CNN, however it does not provide true structured reasoning.

\paragraph{Parameter Learning in CRFs} FullCRFs \cite{DBLP:journals/corr/abs-1210-5644} rely on hand-crafted features for the pairwise (Gaussian) kernels. In the original publication \cite{DBLP:journals/corr/abs-1210-5644} Kr{\"a}henb{\"u}hl and Koltun optimized the remaining parameters with a combination of expectation maximization and grid-search. In a follow-up work \cite{krahenbuhl2013parameter} they proposed to use gradient decent. The idea utilizes, that for the message passing the identity $(k_G * Q)' = k_G * Q'$ is valid. This allows them to train all internal CRF parameters, using backpropagation, without being required to compute gradients with respect to the Gaussian kernel $k_G$. However the features of the Gaussian kernel cannot be learned with such an approach. CRFasRNN \cite{zheng2015conditional} uses the same ideas to implement joint CRF and CNN training. Like \cite{krahenbuhl2013parameter} this approach requires hand-crafted pairwise (Gaussian) features.

Quadratic optimization \cite{chandra2016fast,vemulapalli2016gaussian} has been proposed to learn the Gaussian features of FullCRFs. These approaches however do not fit well into many deep learning pipelines. Another way of learning the pairwise features is piecewise training \cite{lin2016efficient}. An additional advantage of this method is, that it avoids repeated CRF inference, speeding up the training considerably. This approach is however of an approximate nature and inference speed is still very slow.

\paragraph{Inference speed of CRFs} In order to circumvent the issue of very long training and inference times, some CRF based pipelines produce an output which is down-sampled by a factor of $8 \times 8$ \cite{chandra2016fast, lin2016efficient}. This speeds up the inference considerably. However this harms their predictive capabilities. Deep learning based semantic segmentation pipelines perform best when they are challenged to produce a full-resolution prediction \cite{long2015fully,yu2015multi,chen2017rethinking}. To the best of our knowledge, no significant progress in inference speed has been made since the introduction of FullCRFs \cite{DBLP:journals/corr/abs-1210-5644}.


\section{Fully Connected CRFs}


In the context of semantic segmentation most CRF based approaches are based on the Fully Connected CRF (FullCRF) model \cite{DBLP:journals/corr/abs-1210-5644}. Consider an input image $I$ consisting of $n$ pixels and a segmentation task with $k$ classes. A segmentation of $I$ is then modelled as a random field $\mathbf{X} = \{ X_1, \dots, X_n\}$, where each random variable $X_i$ takes values in $\{1, \dots, k\}$, i.e. the label of pixel $i$. Solving $\argmax_X P(X|I)$ then leads to a segmentation $X$ of the input image $I$. $P(X|I)$ is modelled as a CRF over the Gibbs distribution:

\begin{equation}
P(X = \hat x | \tilde I = I ) = \frac{1}{Z(I)} exp(-E(\hat x | I))
\end{equation}

where the energy function $E(\hat x | I)$ is given by 

\begin{equation}
E(\hat x | I) = \sum_{i \leq N} \psi_u (\hat x_i| I ) + \sum_{i \not = j \leq N} \psi_p (\hat x_i, \hat x_j | I).
\end{equation}

The function $\psi_u (x_i | I)$ is called unary potential. The unary itself can be considered a segmentation of the image and any segmentation pipeline can be used to predict the unary. In practise most modern approaches \cite{chen2018deeplab,DBLP:journals/corr/SchwingU15,zheng2015conditional} utilize CNNs to computer the unary. 


The function $\psi_p (x_i, x_j | I)$ is the pairwise potential. It accounts for the joint distribution of pixels $i, j$. It allows us to explicitly model interactions between pixels, such as pixels with similar colour are likely the same class. In FullCRFs $\psi_p$ is defined as weighted sum of Gaussian kernels $k_G^{(1)} \dots k_G^{(1)}$:
\begin{equation}
\psi_p (x_i, x_j | I) := \mu(x_i, x_j) \sum_{m=1}^M w^{(m)} k_G^{(m)} (f_i^I, f_j^I),
\end{equation}
where $w^{(m)}$ are learnable parameters. The feature vectors $f_i^I$ can be chosen arbitrary and may depend on the input Image $I$. The function $\mu(x_i, x_j)$ is the compatibility transformation, which only depends on the labels $x_i$ and $x_j$, but not on the image $I$.

A very widely used compatibility function \cite{DBLP:journals/corr/abs-1210-5644,chen2018deeplab,zheng2015conditional} is the Potts model $\mu(x_i, x_j) = |x_i \not= x_j|$. This model tries to assign pixels with similar features the same prediction. CRFasRNN \cite{zheng2015conditional} proposes to use $1 \times 1$ convolutions as compatibility transformation. Such a function allows the model to learn more structured interactions between predictions.

FullCRFs utilize two Gaussian kernels with hand-crafted features. The appearance kernel $k_\alpha$ uses the raw colour values $I_j$ and $I_i$ as features. The smoothness kernel is based on the spatial coordinates $p_i$ and $p_j$. The entire pairwise potential is then given as:

\begin{equation}
k(f_i^I, f_j^I) := w^{(1)} exp\left(-\frac{|p_i - p_j|^2}{2 \theta^2_\alpha} - \frac{|I_i - I_j|^2}{2 \theta^2_\beta} \right) + w^{(2)} exp\left(-\frac{|p_i - p_j|^2}{2 \theta^2_\gamma} \right),
\end{equation}

where $w^{(1)}$, $w^{(2)}$, as well as $\theta_\alpha$, $\theta_\beta$ and $\theta_\gamma$ are the only learnable parameters of the model. Most CRF based segmentation approaches \cite{chen2018deeplab,zheng2015conditional,DBLP:journals/corr/SchwingU15} utilize the very same handcrafted pairwise potentials. CRFs are notoriously hard to optimize utilizing hand-crafted features circumvents this problem.



\subsection{Mean Field Inference}

Inference in FullCRFs is achieved using the mean field algorithm (see \Cref{alg:mean_field_conv}). All steps of \cref{alg:mean_field_conv}, other then the message passing, are highly parallelized and can be implemented easily and efficiently on GPUs using standard deep learning libraries. (For details see \cite{zheng2015conditional}).

\begin{algorithm}
  \caption{Mean field approximation in convolutional connected CRFs}
  \label{alg:mean_field_conv}
  \begin{algorithmic}[1]
  \State Initialize:   \Comment{$\tilde Q_i \leftarrow \frac{1}{Z_i}exp{(- \psi_u (x_i | I) }) \; \text{"softmax"}$} 
  \While{not converged}
  \State $\tilde Q_i(l) \leftarrow \sum_{i \not= j} w^{(m)} k_G^{(m)} (f_i^I, f_j^I) \tilde Q_i(l)$ \Comment{Message Passing}
  \State $\tilde Q_i(x_i) \leftarrow \sum_{l' \in L} \mu(x_i, l') \tilde Q_i(l)$ \Comment{Compatibility Transformation}
  \State $\tilde Q_i(x_i) \leftarrow  \psi_u (x_i | I ) + \tilde Q_i(x_i)$ \Comment{Adding Unary Potentials}

  \State $\tilde Q_i(x_i) \leftarrow \text{normalize}(\tilde Q_i(x_i))$ \Comment{e.g. softmax}
  \EndWhile
  \end{algorithmic}
\end{algorithm}

The message passing however is the bottleneck of the CRF computation. Exact computation is quadratic in the number of pixels and therefore infeasible. Kr{\"a}henb{\"u}hl and Koltun instead proposed to utilize the permutohedral lattice \cite{adams2010fast} approximation, a high-dimensional filtering algorithm. The permutohedral lattice however is based on a complex data structure. While there is a very sophisticated and fast CPU implementation, the permutohedral lattice does not follow the SIMD \cite{nickolls2008scalable} paradigm of efficient GPU computation. In addition, efficient gradient computation of the permutohedral lattice approximation, is also a non-trivial problem. This is the underlying reason why FullCRF based approaches use hand-crafted features.

\section{Convolutional CRFs}

The convolutional CRFs (ConvCRFs) supplement FullCRFs with a conditional independence assumption. We assume that the label distribution of two pixels $i, j$ are conditionally independent, if for the Manhattan distance $d$ holds $d(i,j) > k$. We call the hyperparameter $k$ filter-size. 

This locality assumption is a very strong assumption. It implies that the pairwise potential is zero, for all pixels whose distance exceed $k$. This reduces the complexity of the pairwise potential greatly. The assumption can also be considered valid, given that CNNs are based on local feature processing and are highly successful. This makes the theoretical foundation of ConvCRFs very promising, strong and valid assumptions are the powerhouse of machine learning modelling.

\subsection{Efficient Message Passing in ConvCRFs}

One of the key contribution of this paper is to show that exact message passing is efficient in ConvCRFs. This eliminates the need to use the permutohedral lattice approximation, making highly efficient GPU computation and complete feature learning possible. Towards this goal we reformulate the message passing step to be a convolution with truncated Gaussian kernel and observe that this can be implemented very similar to regular convolutions in CNNs.

Consider an input $P$ with shape $[bs, c, h, w]$ where $bs, c, h, w$ denote batch size, number of classes, input hight and width respectively. For a Gaussian kernel $g$ defined by feature vectors $f_1 \dots f_d$, each of shape $[bs, h, w]$ we define its kernel matrix by

\begin{equation}
k_g [b, dx, dy, x, y] := exp \left( - \sum_{i=1}^d \frac{|f^{(d)}_i[b, x, y]-f^{(d)}_i[b, x-dx, y-dy]|^2}{2 \dot \theta^{2}_i} \right),
\end{equation}

where $\theta_i$ is a learnable parameter. For a set of Gaussian kernels $g_1 \dots g_s$ we define the merged kernel matrix K as $K := \sum_{i=1}^s w_i \cdot g_i$. The result $Q$ of the combined message passing of all $s$ kernels is now given as:

\begin{equation}
Q[b, c, x, y] = \sum_{dx, dy \leq k} K[b, dx, dy, x, y] \cdot P[b, c, x + dx, y+dy].
\end{equation}

This message passing operation is similar to standard 2d-convolutions of CNNs. In our case however, the filter values depend on the spatial dimensions $x$ and $y$. This is similar to locally connected layers \cite{chen2015locally}. Unlike locally connected layers (and unlike 2d-convolutions), our filters are however constant in the channel dimension $c$. One can view our operation as convolution over the dimension $c$ \footnote{Note that the operation refereed to as convolution in the context of NNs is actually known as \emph{cross-correlation} in the signal processing community. The operation we want to implement however is a "proper" 1d-convolution. This operation is related to but not the same as 1d-convolution in NNs.}. 

It is possible to implement our convolution operation by using standard CNN operations only. This however requires the data to be reorganized in GPU memory several times, which is a very slow process. Profiling shows that \SI{90}{\percent} of GPU time is spend for the reorganization of data. We therefore opted to build a native low-level implementation, to gain an additional \num{10}-fold speed up.

Efficient computation of our convolution can be implemented analogously to 2d-convolution (and locally connected layers). The first step is to tile the input $P$ in order to obtain data with shape $[bs, c, k, k, h, w]$. This process is usually referred to as \emph{im2col} and the same as in 2d-convolutions \cite{DBLP:journals/corr/ChetlurWVCTCS14}. 2d-convolutions proceed by applying a batched matrix multiplication over the spatial dimension. We replace this step with a batched dot-product over the channel dimension. All other steps are the same.

\subsection{Additional implementation details}

For the sake of comparability we use the same design choices as FullCRFs in our baseline ConvCRF implementation. In particular, we use softmax  normalization, the Potts model as well as the same hand-crafted gaussian features as \cite{DBLP:journals/corr/abs-1210-5644,chen2018deeplab,zheng2015conditional}. Analogous to  \cite{DBLP:journals/corr/abs-1210-5644} we apply gaussian blur to the pairwise kernels. This leads to an increase of the effective filter size by a factor of $4$.

In additional experiments we investigate the capability of our CRFs to learn Gaussian features. Towards this goal we replace the input features $p_i$ of the smoothness kernel with learnable variables. Those variables are initialized to the same values as the  hand-crafted version, but are adjusted as part of the training process. We also implement a learnable compatibility transformation using $1 \times 1$ convolution, following the ideas of \cite{zheng2015conditional}. 

\section{Experimental Evaluation}


\paragraph{Dataset:} We evaluate our method on the challenging PASCAL VOC 2012 \cite{pascal-voc-2012} image dataset. Following the literature \cite{long2015fully,DBLP:journals/corr/WuSH16e,chen2018deeplab,zhao2017pyramid} we use the additional annotation provided by \cite{hariharan2011semantic} resulting in \num{10582} labelled images for training. Out of those images we hold back \num{200} images to fine-tune the internal CRF parameters and use the remaining \num{10382} to train the unary CNN. We report our results on the \num{1464} images of the official validation set.

\paragraph{Unary:} \label{par:unary} We train a ResNet101 \cite{he2016deep} to compute the unary potentials. We use the ResNet101 implementation provided by the PyTorch \cite{paszke2017automatic} repository. A simple FCN \cite{long2015fully} is added on top of the ResNet to decode the CNN features and obtain valid segmentation predictions. The network is initialized using ImageNet Classification weights \cite{imagenet_cvpr09} and then trained on Pascal VOC data directly. Unlike many other projects, we do not train the network on large segmentation datasets such as MS COCO \cite{lin2014microsoft}, but only use the images provided by the PASCAL VOC 2012 benchmark.

The CNN is trained for $200$ epochs using a batch size of $16$ and the adam optimizer \cite{kingma2014adam}. The initial learning rate is set to \num{5e-5} and polynomially decreased \cite{liu2015parsenet,chen2018deeplab} by multiplying the initial learning rate with $((1-\frac{step}{max\_steps})^{0.9})^2$. An $L_2$ weight decay with factor \num{5e-4} is applied to all kernel weights and 2d-Dropout \cite{tompson2015efficient} with rate $0.5$ is used on top of the final convolutional layer.  The same hyperparamters are also used for the end-to-end training.

The following data augmentation methods are applied: Random horizontal flip, random rotation (\ang{\pm 10}) and random resize with a factor in (0.5, 2). In addition the image colours are jittered using random brightness, random contrast, random saturation and random hue. All random numbers are generated using a truncated normal distribution. The trained model achieves validation mIoU of \SI{71.23}{\percent} and a train mIoU of \SI{91.84}{\percent}.

\paragraph{CRF:} Following the literature \cite{DBLP:journals/corr/abs-1210-5644,chen2018deeplab,zheng2015conditional,lin2016efficient}, the mean-field inference of the CRF is computed for $5$ iterations in all experiments. For training, those iterations are unrolled.

\begin{table}
\centering
\begin{tabular}{l | r r r r r r}
\toprule
 Method & Unary & FullCRF & Conv5 & Conv7 & Conv11 & Conv13 \\
\midrule
Speed [ms]    & 68  &  647   & 7 & 13  &  26 & 34 \\
Accuracy [\%] & 86.60 & 94.79  & 97.13 & 97.13 & 98.97 & 98.99 \\
mIoU     [\%] & 51.87 & 84.37  & 90.90 & 92.98 & 93.74 & 93.89 \\ 
\bottomrule
\end{tabular}
\caption{Performance comparison of CRFs on synthetic data. The speed tests have been done on a Nvidia GeFore GTX 1080 Ti GPU. All CRFs are applied on full resolution Pascal VOC images. Conv7 denotes a ConvCRF with filter size 7.}
\label{tab:synthetic_results}
\end{table}


\subsection{ConvCRFs on synthetic data}

\label{sec:syn}


To show the capabilities of Convolutional CRFs we first evaluate their performance on a synthetic task. We use the PASCAL VOC \cite{pascal-voc-2012} dataset as a basis, but augment the ground-truth with the goal to simulate prediction errors. The noised labels are used as unary potentials for the CRF, the CRF is then challenged to denoise the predictions. The output of the CRF is then compared to the original label of the Pascal VOC dataset.

\begin{figure}[tp]
\begin{tabular}{ccccc}
\hspace{-0.2cm}\bmvaHangBox{\includegraphics[width=2.6cm]{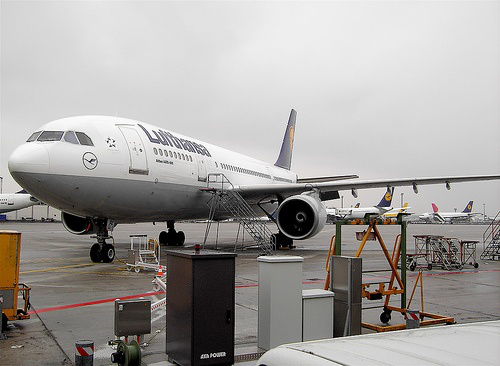}\hspace{-0.45cm}}&
\bmvaHangBox{\includegraphics[width=2.6cm]{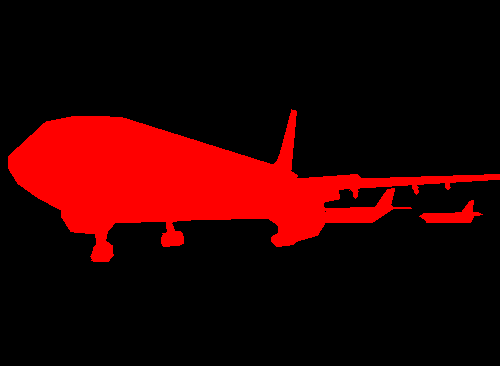}\hspace{-0.45cm}}&
\bmvaHangBox{\includegraphics[width=2.6cm]{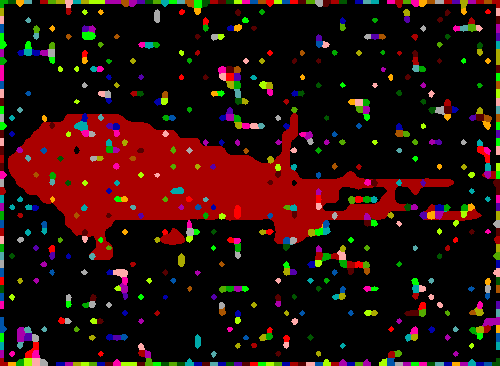}\hspace{-0.45cm}}&
\bmvaHangBox{\includegraphics[width=2.6cm]{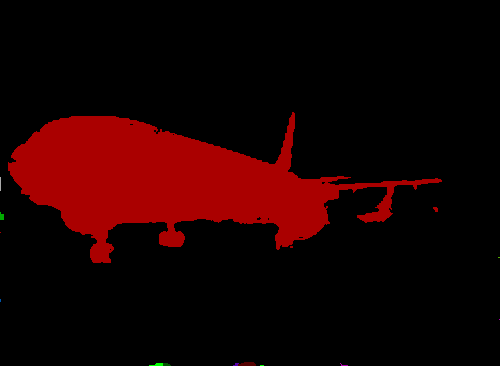}\hspace{-0.5cm}}&
\bmvaHangBox{\includegraphics[width=2.6cm]{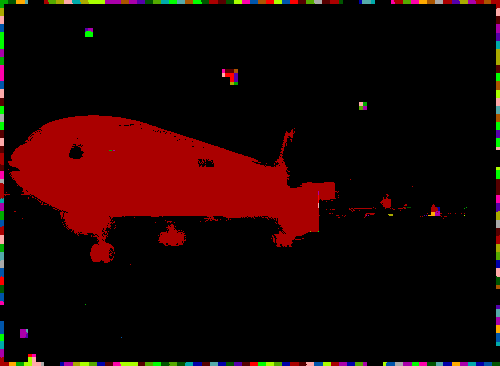}\hspace{-0.45cm}}\\

\hspace{-0.2cm}\bmvaHangBox{\includegraphics[width=2.6cm]{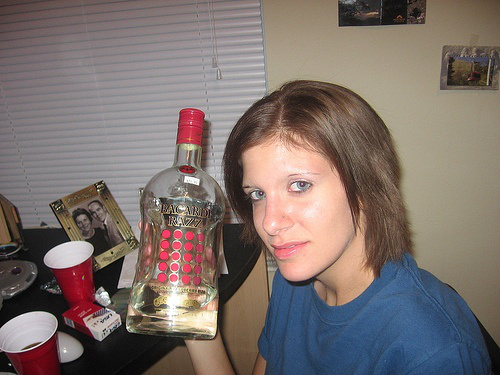}\hspace{-0.45cm}}&
\bmvaHangBox{\includegraphics[width=2.6cm]{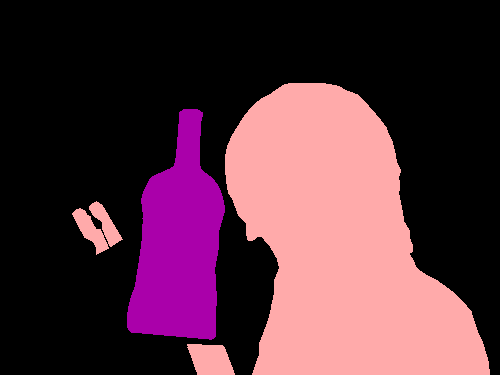}\hspace{-0.45cm}}&
\bmvaHangBox{\includegraphics[width=2.6cm]{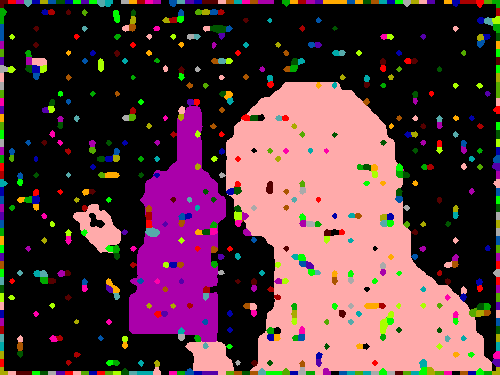}\hspace{-0.45cm}}&
\bmvaHangBox{\includegraphics[width=2.6cm]{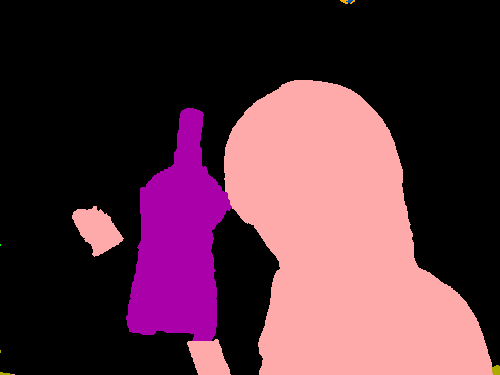}\hspace{-0.5cm}}&
\bmvaHangBox{\includegraphics[width=2.6cm]{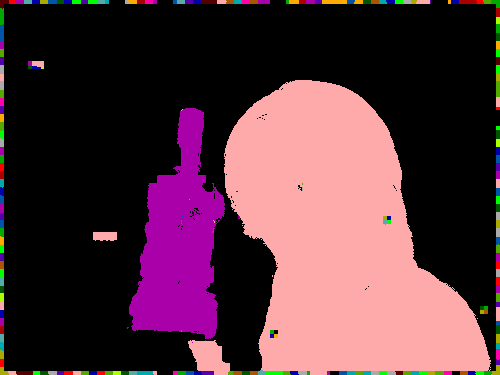}\hspace{-0.45cm}}\\

\hspace{-0.2cm}\bmvaHangBox{\includegraphics[width=2.6cm]{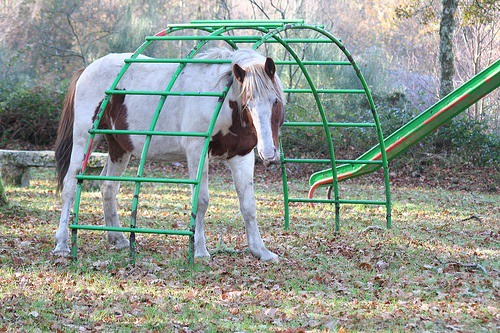}\hspace{-0.45cm}}&
\bmvaHangBox{\includegraphics[width=2.6cm]{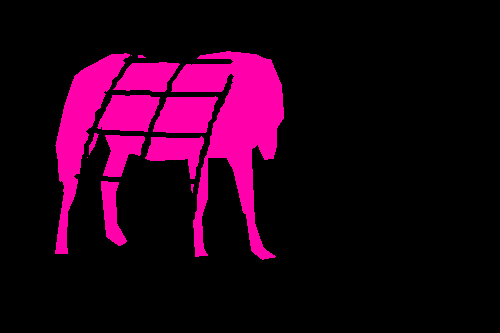}\hspace{-0.45cm}}&
\bmvaHangBox{\includegraphics[width=2.6cm]{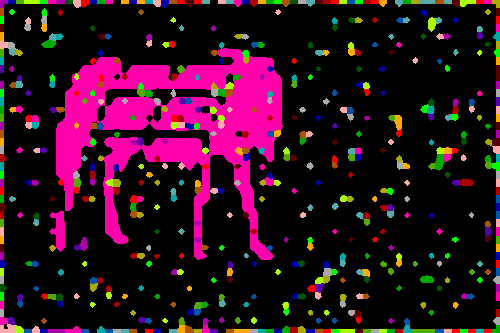}\hspace{-0.45cm}}&
\bmvaHangBox{\includegraphics[width=2.6cm]{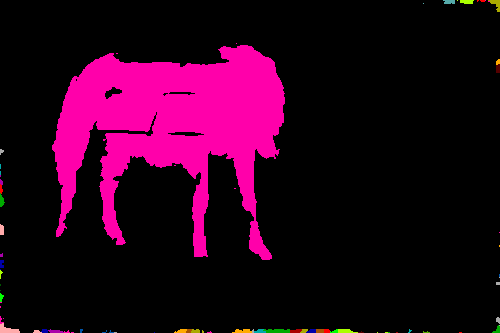}\hspace{-0.5cm}}&
\bmvaHangBox{\includegraphics[width=2.6cm]{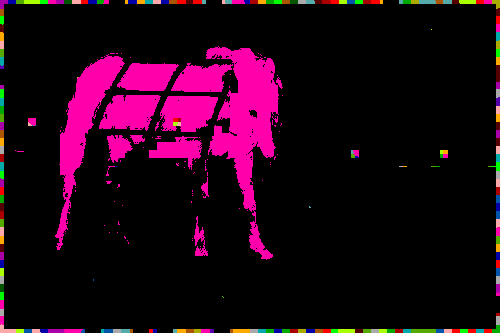}\hspace{-0.45cm}}\\

\hspace{-0.2cm}\bmvaHangBox{\includegraphics[width=2.6cm]{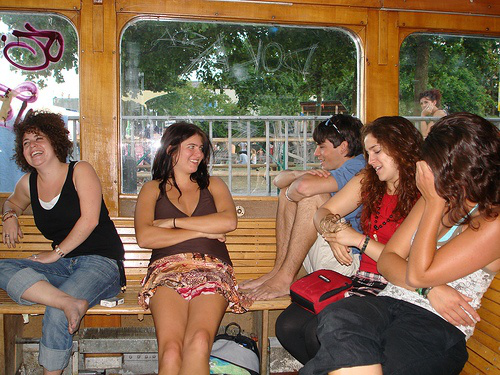}\hspace{-0.45cm}}&
\bmvaHangBox{\includegraphics[width=2.6cm]{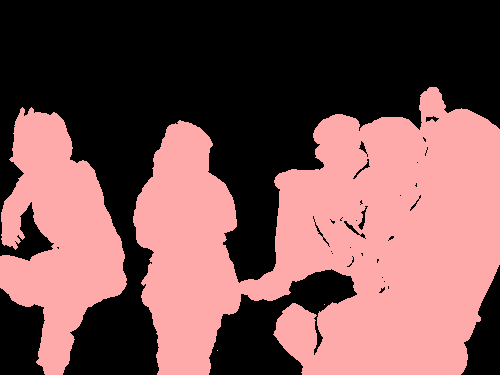}\hspace{-0.45cm}}&
\bmvaHangBox{\includegraphics[width=2.6cm]{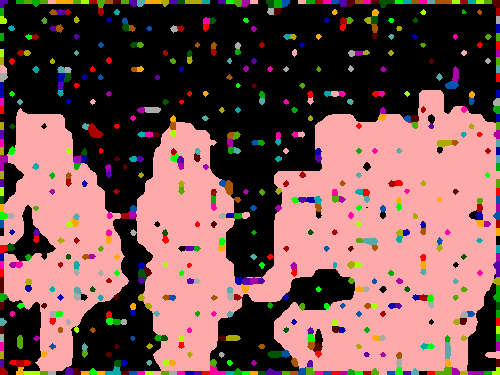}\hspace{-0.45cm}}&
\bmvaHangBox{\includegraphics[width=2.6cm]{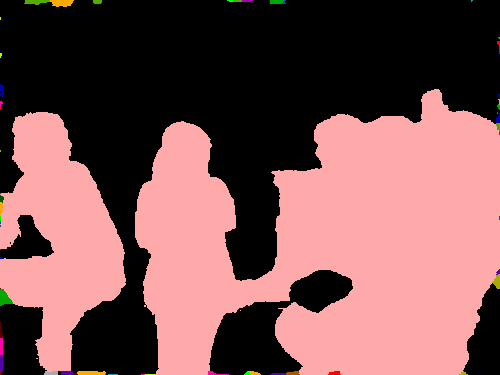}\hspace{-0.5cm}}&
\bmvaHangBox{\includegraphics[width=2.6cm]{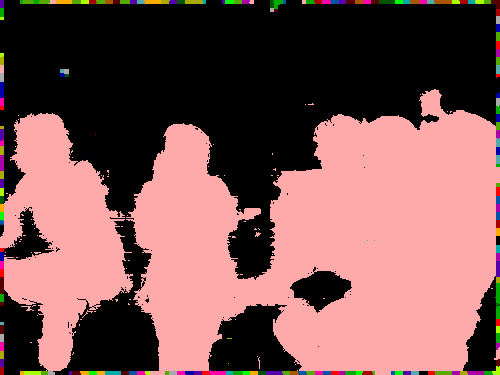}\hspace{-0.45cm}}\\

\hspace{-0.2cm}\bmvaHangBox{\includegraphics[width=2.6cm]{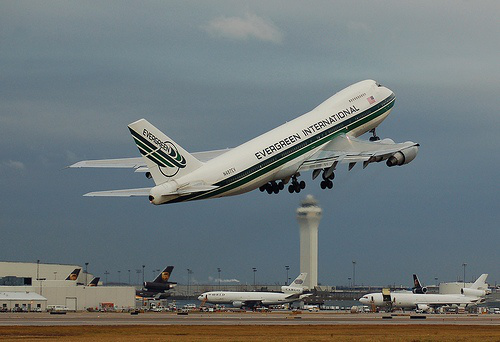}\hspace{-0.45cm}}&
\bmvaHangBox{\includegraphics[width=2.6cm]{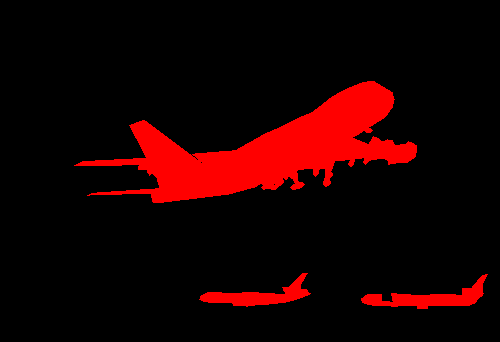}\hspace{-0.45cm}}&
\bmvaHangBox{\includegraphics[width=2.6cm]{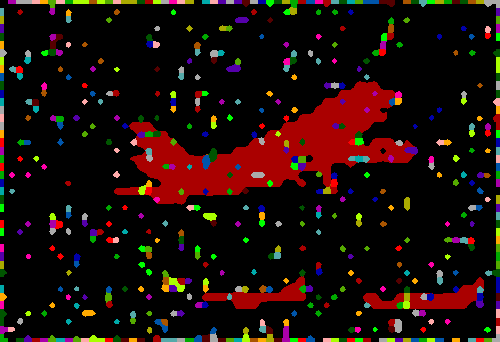}\hspace{-0.45cm}}&
\bmvaHangBox{\includegraphics[width=2.6cm]{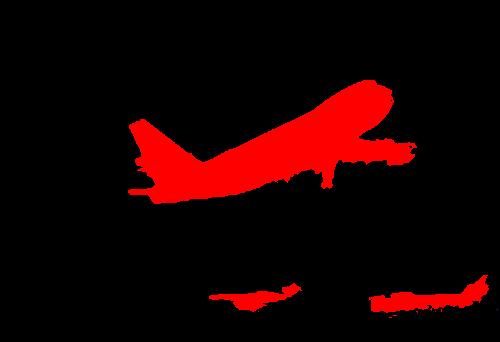}\hspace{-0.5cm}}&
\bmvaHangBox{\includegraphics[width=2.6cm]{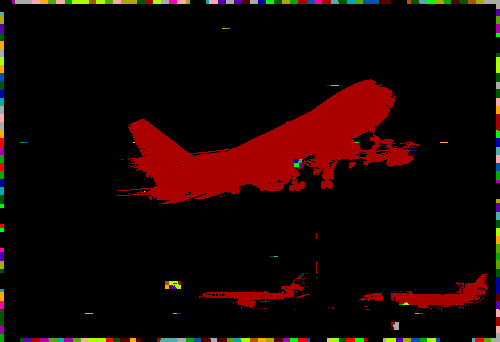}\hspace{-0.45cm}}\\

\footnotesize{(a) Image} &\footnotesize{(b) Label}&\footnotesize{(c) Noised}&\footnotesize{(d) ConvCRF11}&\footnotesize{(e) FullCRF}
\end{tabular}
\caption{Visualization of the synthetic task. Especially in the last example, the artefacts from the permutohedral lattice approximation can clearly be seen at object boundaries.}
\label{fig:vis_results}
\end{figure}

Towards the goal of creating a relevant task, the following augmentation procedure is used: First the ground-truth is down-sampled by a factor of 8. Then, in low-resolution space predictions are randomly flipped and the result is up-sampled to the original resolution again. This process simulates inaccuracies as a result of the low-resolution feature processing of CNNs as well as prediction errors similar to the checkerboard issue found in deconvolution based segmentation networks \cite{DBLP:journals/corr/GaoYWJ17,odena2016deconvolution}. Some examples of the augmented ground-truth are shown in \Cref{fig:vis_results}.

In our first experiment we compare FullCRFs and ConvCRFs using the exact same parameters. To do this we utilize the hand-crafted Gaussian features. The remaining five parameters (namely $w^{(1)}$, $w^{(2)}$, as well as $\theta_\alpha$, $\theta_\beta$ and $\theta_\gamma$) are initialized to the default values proposed in \cite{DBLP:journals/corr/abs-1210-5644,dcrfCode}. Note that this gives FullCRFs a natural advantage. The performance of CRFs however is very robust with respect to these five parameters \cite{DBLP:journals/corr/abs-1210-5644}.


The results of our first experiment are given in \Cref{tab:synthetic_results}. It can be seen that ConvCRFs outperform FullCRFs significantly. This shows that ConvCRFs are structurally superior to FullCRFs. The better performance of ConvCRFs with the same parameters can be explained by our exact message passing, which avoids the approximation errors compared of the permutohedral lattice approximation. We provide a visual comparison in \Cref{fig:vis_results} where ConfCRF clearly provide higher quality output. The FullCRF output shows approximation artefacts at the boundary of objects. In addition we note that ConvCRFs are faster by two orders of magnitude, making them favourable in almost every use case.

\subsection{Decoupled training of ConvCRFs}

In this section we discuss our experiments on Pascal VOC data using a two stage training strategy. First the unary CNN model is trained to perform semantic segmentation on the Pascal VOC data. Those parameters are then fixed and in the second stage the internal CRF parameters are optimized with respect to the CNN predictions. The same unary predictions are used across all experiments, to reduce variants between runs.

\begin{figure}
    \centering

\begin{tabular}{cc}
\hspace{-0.2cm}\bmvaHangBox{\includegraphics[width=0.5\textwidth]{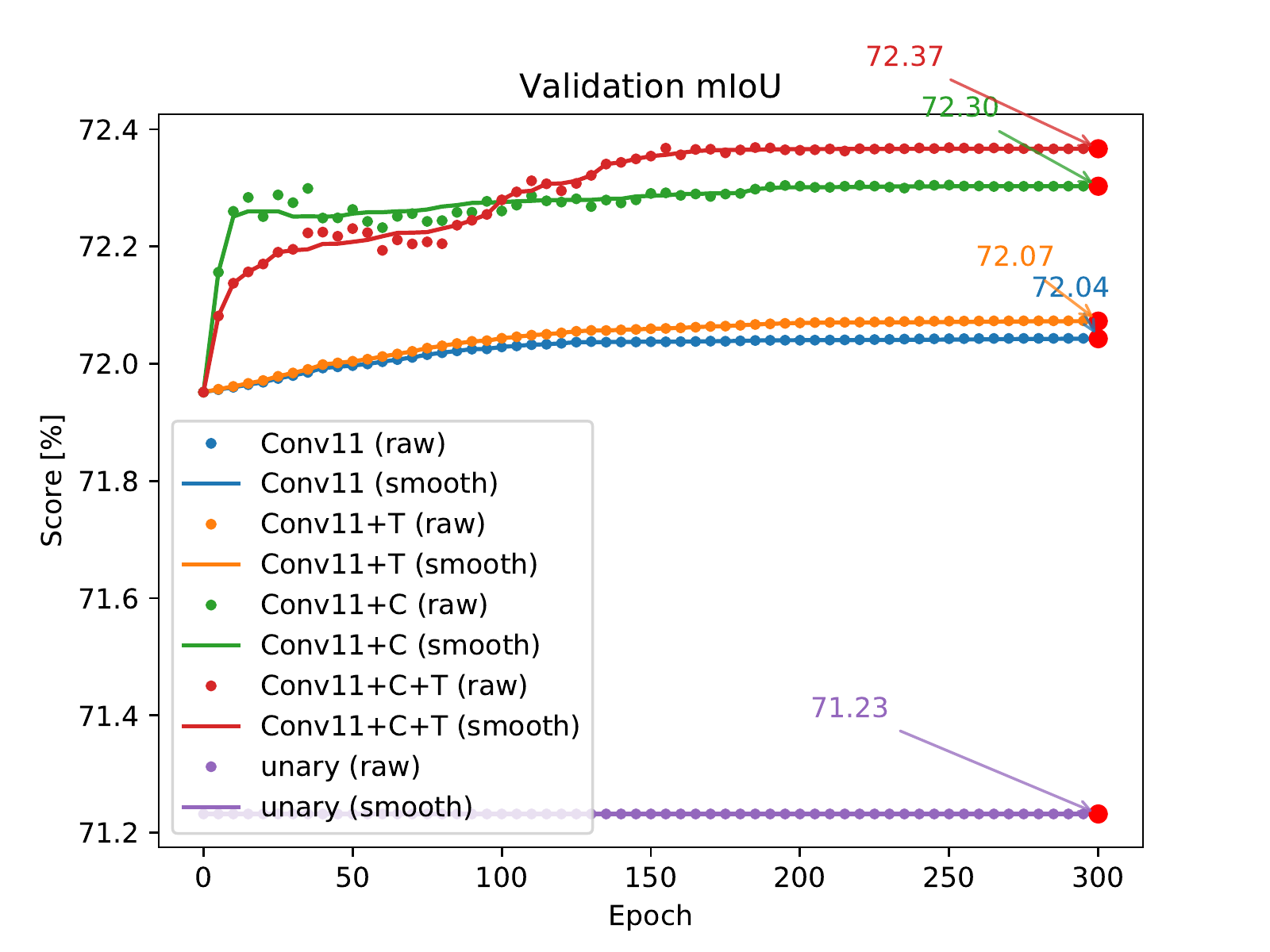}\hspace{-0.45cm}}&
\bmvaHangBox{\includegraphics[width=0.5\textwidth]{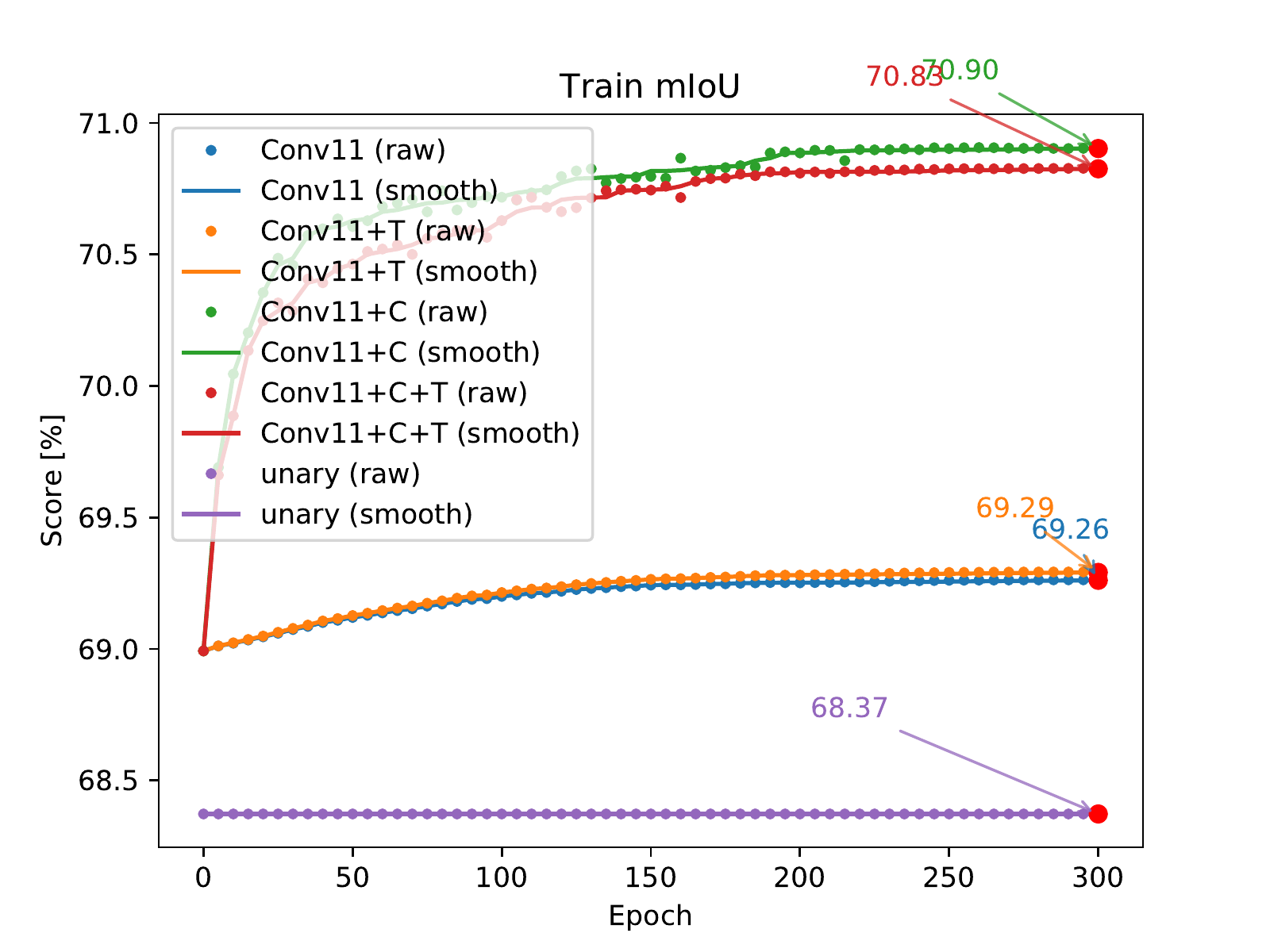}\hspace{-0.45cm}}\\

\footnotesize{(a) Validation mIoU over time} &\footnotesize{(b) Train mIoU over time}
\end{tabular}
    \caption{Training and validation mIoU over time. +C uses convolutions as compatibility transformation and +T learns the features for the smoothness kernel.}
    \label{fig:learning_curves} 
\end{figure}

Decoupled training has various merits compared to an end-to-end pipeline. Firstly it is very flexible. A standalone CRF training can be applied on top of any segmentation approach. The unary predictions are treated as a black-box input for the CRF training. In practice this means that the two training stages do not need to interface at all, making fast prototyping very easy. Additionally decoupled training keeps the system interpretable. Lastly, piecewise training effectively tackles the vanishing gradient problem \cite{bengio1994learning}, which is still an issue in CNN based segmentation approaches \cite{DBLP:journals/corr/WuSH16e}. This leads to overall faster, more robust and reliable training.

For our experiments we train the CRF models on the \num{200} held-out images from the training set and evaluate the CRF performance on the \num{1464} images of the official Pascal VOC dataset. We compare the performance of the ConvCRF with filter size \num{11} to the unary baseline results as well as a FullCRF trained following the methodology of DeepLab \cite{chen2018deeplab}. More context about the training methodology can be found in the supplementary material.

\begin{figure}[tp]
\begin{tabular}{ccccc}
\hspace{-0.2cm}\bmvaHangBox{\includegraphics[width=2.6cm]{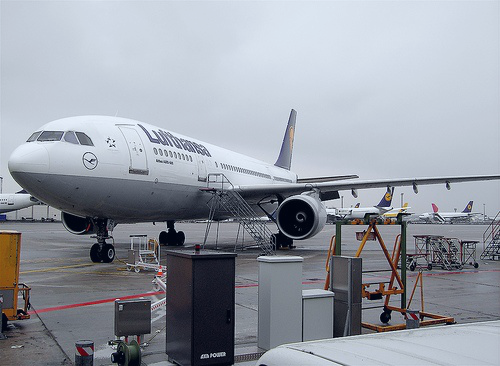}\hspace{-0.45cm}}&
\bmvaHangBox{\includegraphics[width=2.6cm]{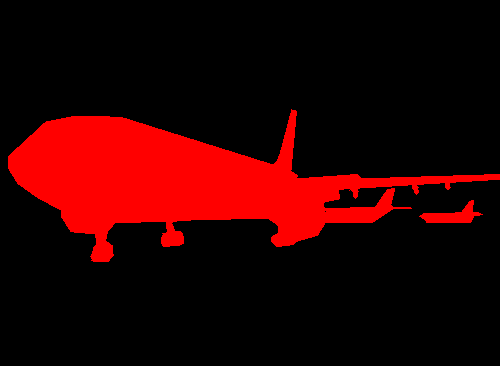}\hspace{-0.45cm}}&
\bmvaHangBox{\includegraphics[width=2.6cm]{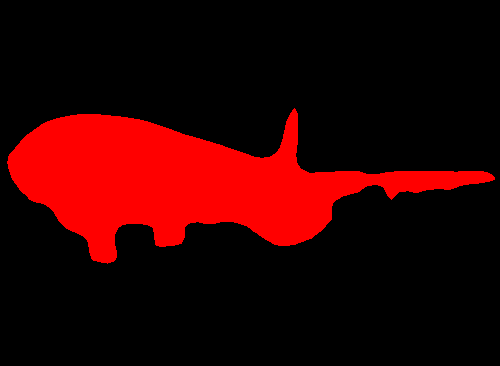}\hspace{-0.45cm}}&
\bmvaHangBox{\includegraphics[width=2.6cm]{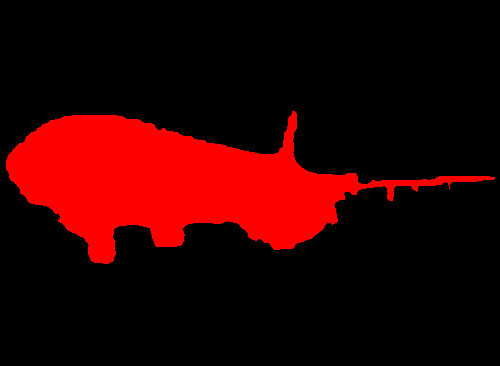}\hspace{-0.45cm}}&
\bmvaHangBox{\includegraphics[width=2.6cm]{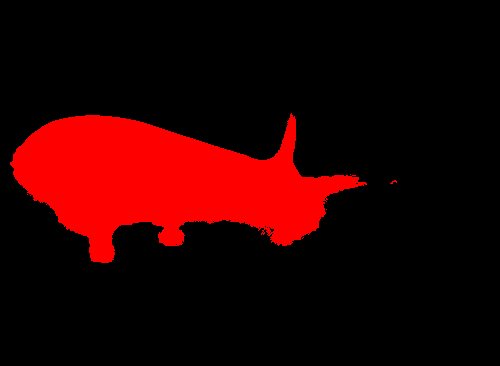}\hspace{-0.5cm}}\\

\hspace{-0.2cm}\bmvaHangBox{\includegraphics[width=2.6cm]{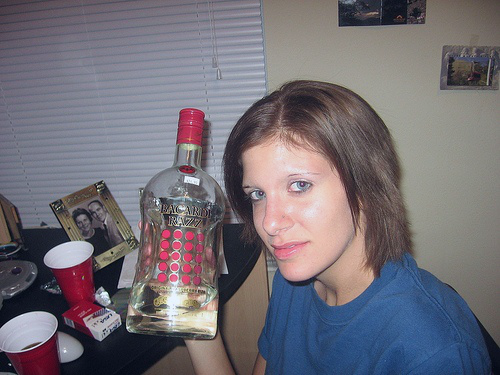}\hspace{-0.45cm}}&
\bmvaHangBox{\includegraphics[width=2.6cm]{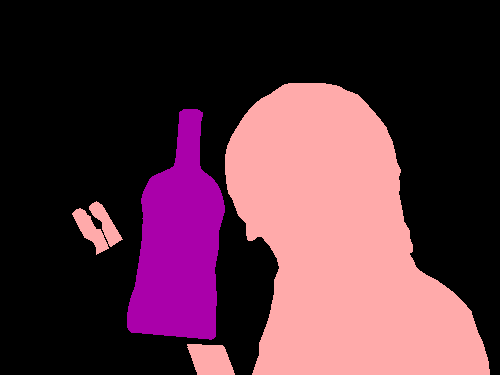}\hspace{-0.45cm}}&
\bmvaHangBox{\includegraphics[width=2.6cm]{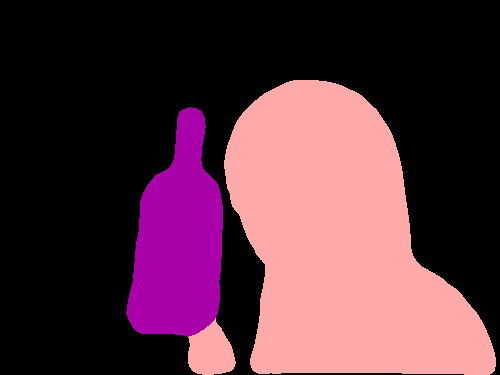}\hspace{-0.45cm}}&
\bmvaHangBox{\includegraphics[width=2.6cm]{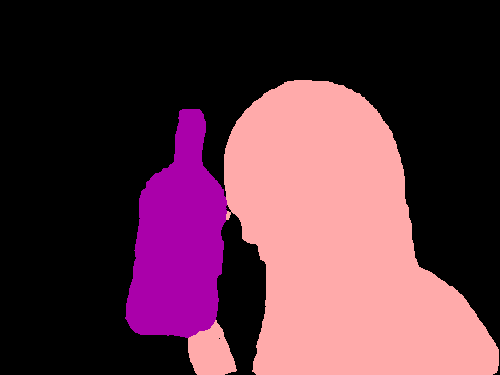}\hspace{-0.45cm}}&
\bmvaHangBox{\includegraphics[width=2.6cm]{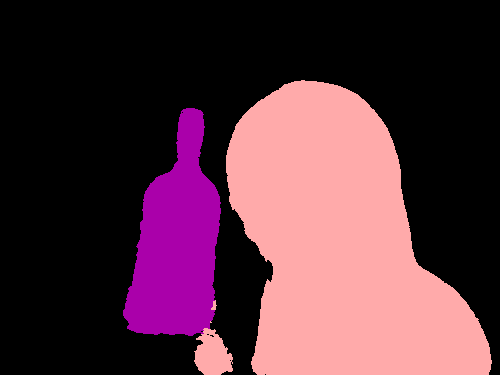}\hspace{-0.5cm}}\\

\hspace{-0.2cm}\bmvaHangBox{\includegraphics[width=2.6cm]{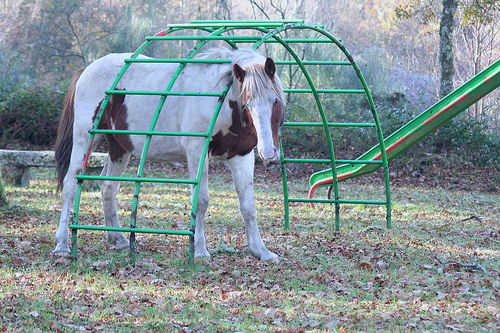}\hspace{-0.45cm}}&
\bmvaHangBox{\includegraphics[width=2.6cm]{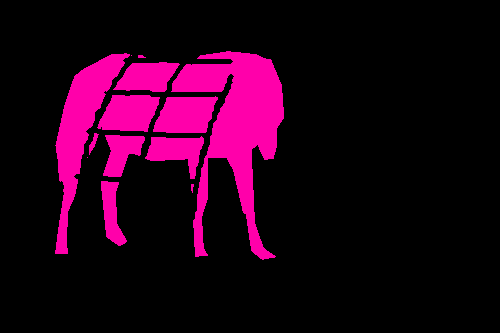}\hspace{-0.45cm}}&
\bmvaHangBox{\includegraphics[width=2.6cm]{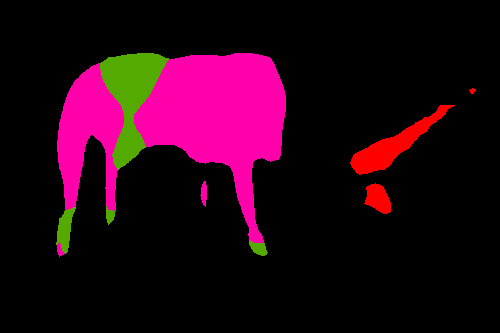}\hspace{-0.45cm}}&
\bmvaHangBox{\includegraphics[width=2.6cm]{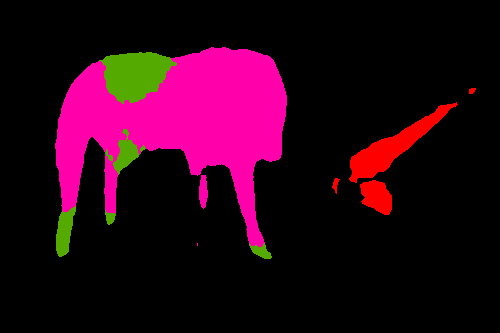}\hspace{-0.45cm}}&
\bmvaHangBox{\includegraphics[width=2.6cm]{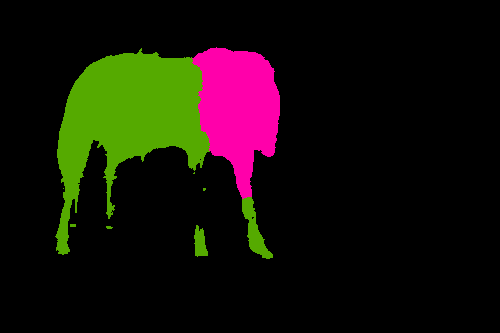}\hspace{-0.5cm}}\\

\hspace{-0.2cm}\bmvaHangBox{\includegraphics[width=2.6cm]{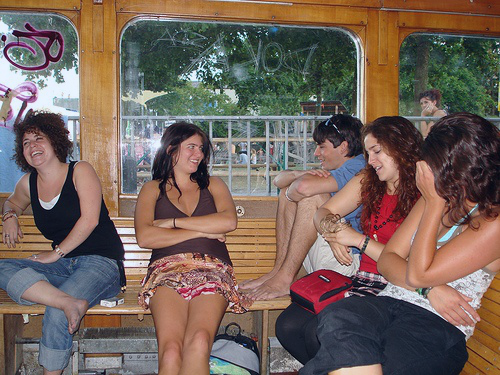}\hspace{-0.45cm}}&
\bmvaHangBox{\includegraphics[width=2.6cm]{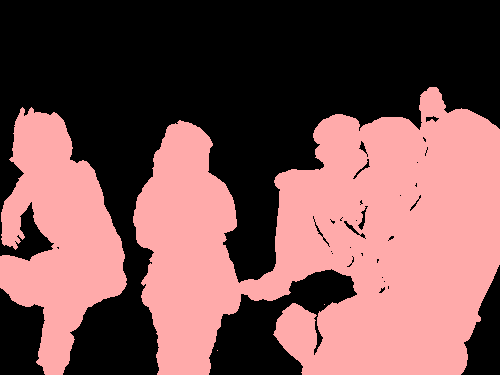}\hspace{-0.45cm}}&
\bmvaHangBox{\includegraphics[width=2.6cm]{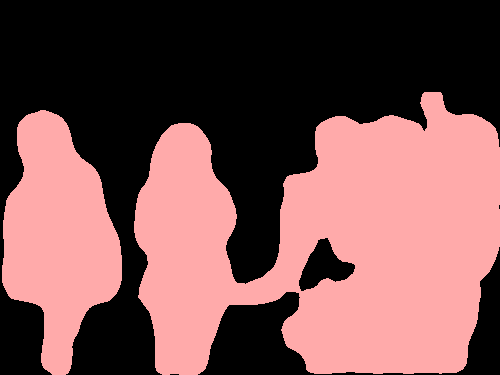}\hspace{-0.45cm}}&
\bmvaHangBox{\includegraphics[width=2.6cm]{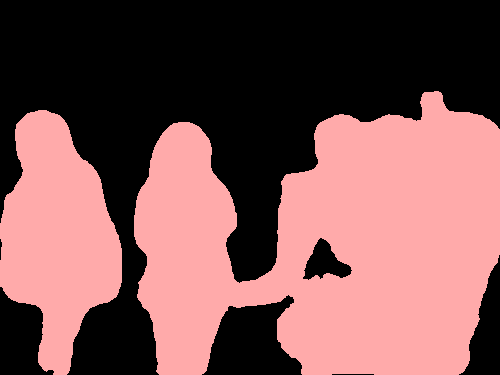}\hspace{-0.45cm}}&
\bmvaHangBox{\includegraphics[width=2.6cm]{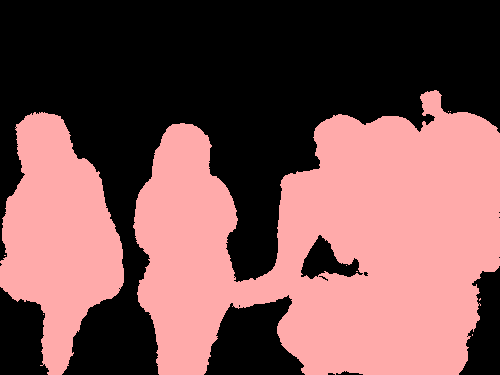}\hspace{-0.5cm}}\\

\hspace{-0.2cm}\bmvaHangBox{\includegraphics[width=2.6cm]{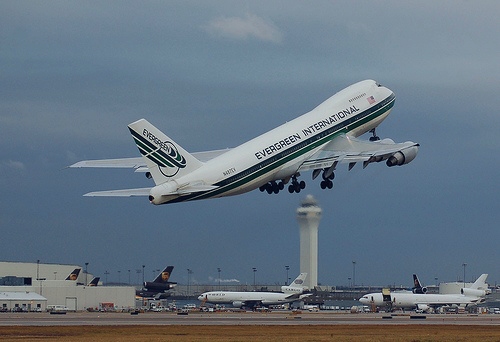}\hspace{-0.45cm}}&
\bmvaHangBox{\includegraphics[width=2.6cm]{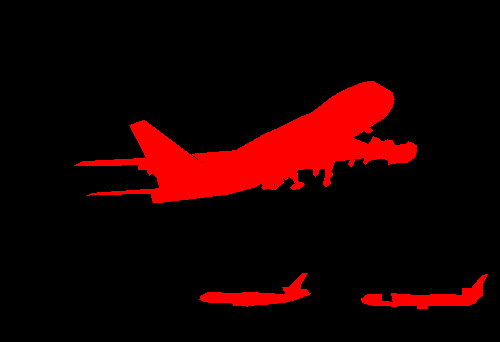}\hspace{-0.45cm}}&
\bmvaHangBox{\includegraphics[width=2.6cm]{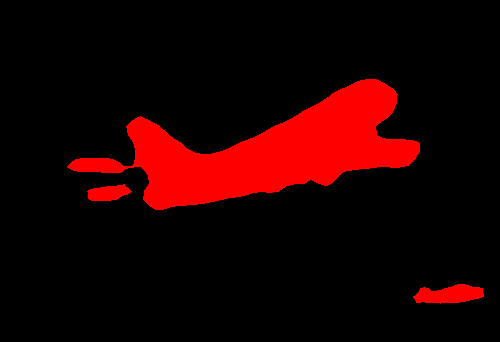}\hspace{-0.45cm}}&
\bmvaHangBox{\includegraphics[width=2.6cm]{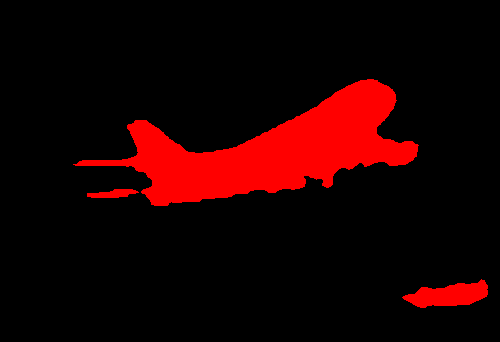}\hspace{-0.45cm}}&
\bmvaHangBox{\includegraphics[width=2.6cm]{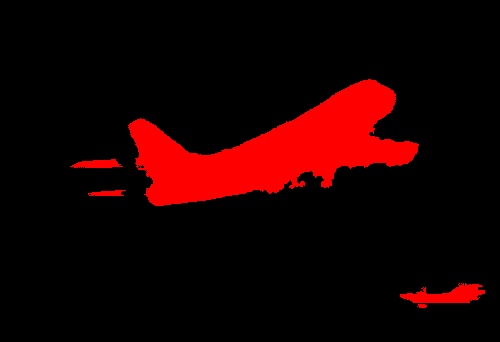}\hspace{-0.5cm}}\\

\footnotesize{(a) Image} &\footnotesize{(b) Label}&\footnotesize{(c) Unary}&\footnotesize{(d) ConvCRF}&\footnotesize{(e) DeepLab-CRF}
\end{tabular}
\caption{Visualization of results on Pascal VOC data using a decoupled training strategy. Examples \num{2} and \num{4} depict failure cases, in which the CRFs are not able to improve the unary.}
\label{fig:voc_results}
\end{figure}

We report our results in \Cref{tab:val_results}. In all experiments, applying CRFs boost the performance considerably. The experiments also confirm the observation of \Cref{sec:syn}, that ConvCRF perform slightly better then FullCRFs. We also observe that the ConvCRF implementation utilizing a learnable compatibility transformation as well as learnable Gaussian features performs best. Model output is visualized in \Cref{fig:voc_results}.

\begin{table}
\centering
\begin{tabular}{l | r r r r r r}
\toprule
 Method & Unary & DeepLab & ConvCRF & Conv+T & Conv+C & \textbf{Conv+CT}\\
\midrule
mIoU     [\%] & 71.23   & 72.02 & 72.04 & 72.07 & 72.30 & \textbf{72.37} \\ 
Accuracy [\%] & 91.84   & 94.01 & 93.99 & 94.01 & 94.01 & \textbf{94.03} \\
train mIoU [\%] & 68.37 & 68.61 & 69.26 & 69.29 & 70.90 & \textbf{70.83} \\ 
\bottomrule
\end{tabular}
\caption{Performance comparison of CRFs on validation data using decoupled training. +C uses convolutions as compatibility transformation and +T learns the Gaussian features. The same unaries were used for all approaches, only the CRF code from DeepLab was utilized.}
\label{tab:val_results}
\end{table}


\subsection{End-to-End learning with ConvCRFs}

In this section we discuss our experiments using an end-to-end learning strategy for ConvCRFs. In end-to-end training the gradients are propagated through the entire pipeline. This allows the CNN and CRF model to co-adapt and therefore to produce the optimum output w.r.t the entire network. The down-side of end-to-end training is that the gradients need to be propagated through five iterations of the mean-field inference, resulting in vanishing gradients \cite{zheng2015conditional}.

We train our network for 250 epochs using a training protocol similar to CRFasRNN \cite{zheng2015conditional}. Zheng et al. propose to first train the unary potential until convergence and then optimizing the CRF and CNN jointly. We achieved best results when limiting the unary-only training to 100 epochs. Afterwards we optimize the CRF and CNN jointly, but introduce an auxiliary unary loss to counterbalance the vanishing gradient problem. In addition, gradient update steps are alternated between unary only and joint gradient update. At the end of each epoch the internal CRF parameters are fine-tuned with the 200 images of the held out training set. During this fine-tuning the CNN parameters are fixed.

The entire training process takes about \num{30} hours using four \num{1080}Ti GPUs in parallel. We believe that the fast training and inference speeds will greatly benefit and ease future research using CRFs. We compare our training protocol to the approach proposed in CRFasRNN \cite{zheng2015conditional} and report the results in \Cref{tab:e2e_results}. The comparison is not entirely fair, additional details regarding the baseline choice as well as some plots of the training process can be found in the supplementary material.

\begin{table}
\centering
\begin{tabular}{l | r r r r r r}
\toprule
 Method & Unary & ConvCRF & CRFasRNN \\
\midrule
mIoU     [\%] & 70.99   & 72.18 & 69.6  \\ 
Accuracy [\%] & 93.76   & 94.04 & 93.03  \\
train mIoU [\%] & 94.90 & 95.31 & 93.25  \\ 
\bottomrule
\end{tabular}
\caption{Performance comparison of end-to-end trained CRFs. CRFasRNN training follows the protocol presented in \cite{zheng2015conditional}, this is further discussed in the supplementary material.}
\label{tab:e2e_results}
\end{table}

\section*{Supplementary Material.}

\subsection*{Decoupled training protocol}

We use a held-out subset of the PASCAL VOC training data to train the internal CRF parameters. We observed that this yields the best results. We believe that this is due to the difference in the unary distribution between training and validate data (71.23 \% vs 91.84 \% mIoU). In the Potts model, the main purpose of the internal CRF parameter is to balance the influence of the unary and pairwise potentials. Thus, the CRF needs an accurate representation how "trustworthy" the unaries are.

In this regard our approach follows the training protocol proposed by DeepLab \cite{chen2018deeplab}. DeepLab uses a grid-search on the validation data utilizing cross-validation. We use a held-out subset of the training data instead.

It is not uncommon for CNN based models to have a very low training error (compared to validation error). We have looked at the training error of several well known segmentation approaches, with publicly available weights. All of those have training scores >\num{90}\% but validation/test scores around 70\%. For additional insights also see the discussion regarding rethinking generalization \cite{zhang2016understanding}. 

\begin{figure}
    \centering

\begin{tabular}{cc}
\hspace{-0.2cm}\bmvaHangBox{\includegraphics[width=0.5\textwidth]{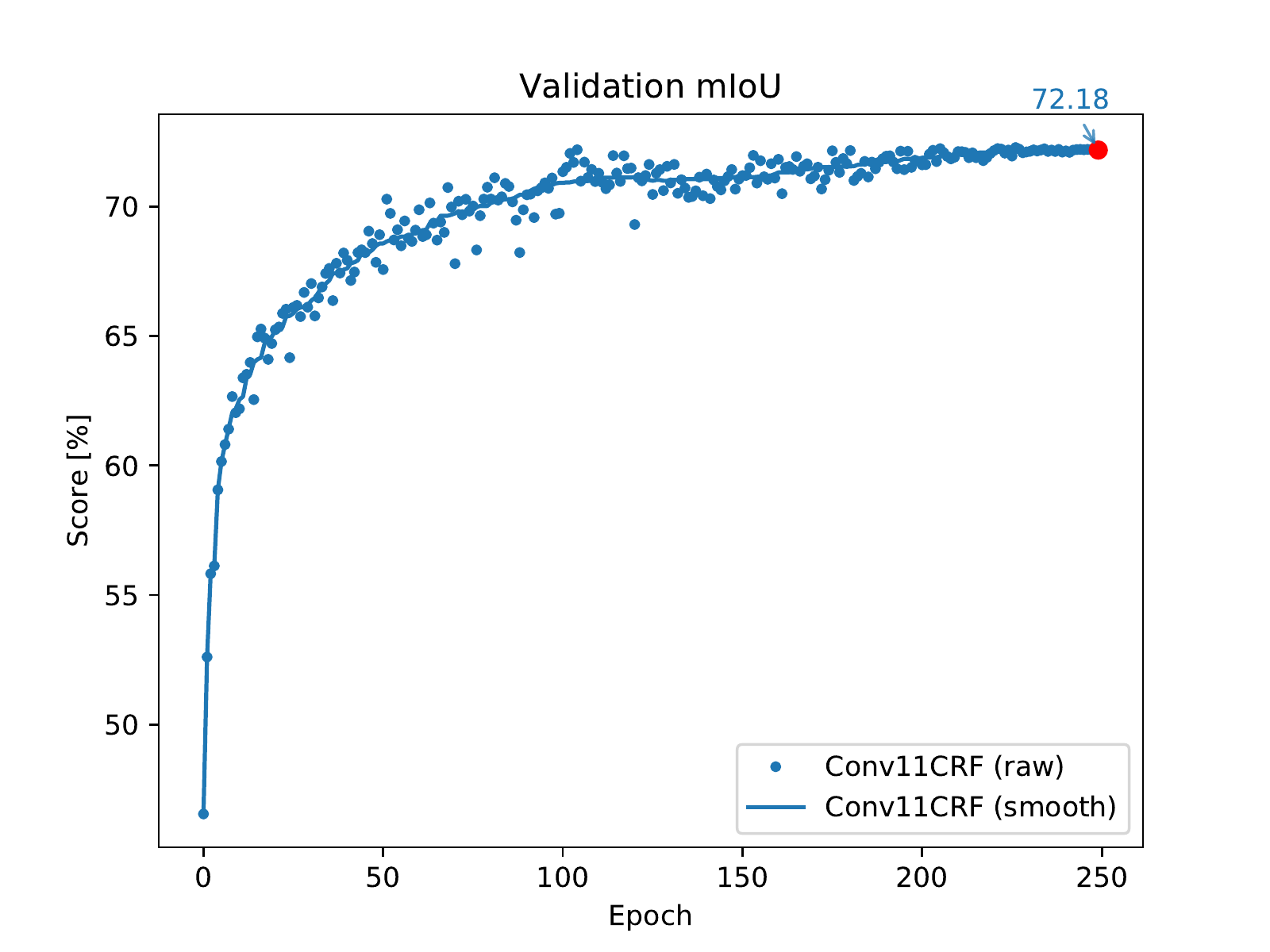}\hspace{-0.45cm}}&
\bmvaHangBox{\includegraphics[width=0.5\textwidth]{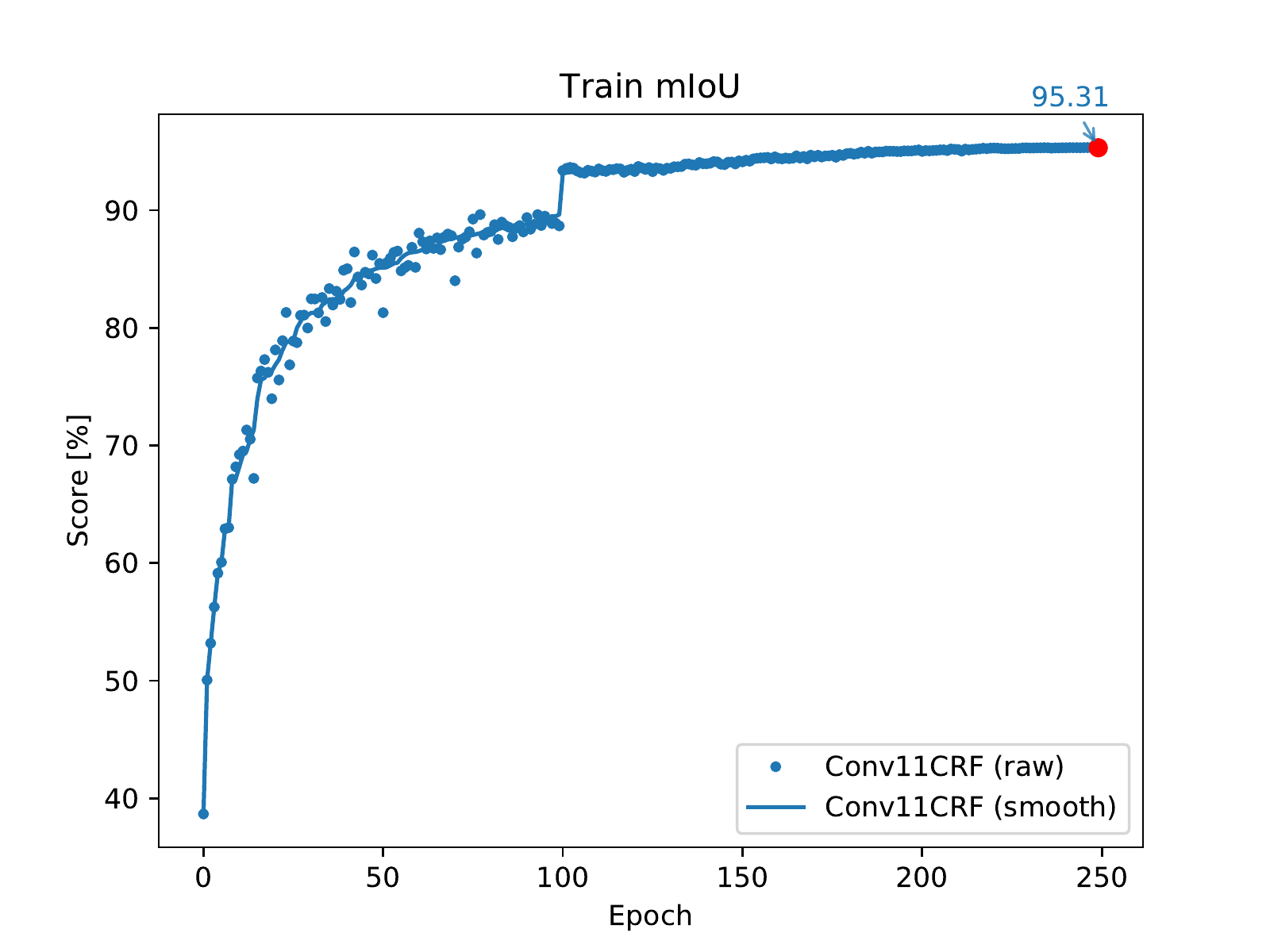}\hspace{-0.45cm}}\\

\footnotesize{(a) Validation mIoU over time} &\footnotesize{(b) Train mIoU over time}
\end{tabular}
    \caption{Training and validation mIoU over time. End-to-end training starts at epoch 100. Note the spike in training mIoU at Epoch 100.}
    \label{fig:learning_end2end} 
\end{figure}

\subsection*{End-to-end training protocol}

We find that starting the joint training before the unary learning converges is crucial to obtain a high validation score. Our unary is trained using a decreasing learning rate. This follows the theory of simulated annealing \cite{van1987simulated} and helps to find relatively good local optima. However when restarting the training after 200 epochs, using a higher learning rate again, the validation score of the unary drops to around $69\%$ and does not recover. Our approach circumvents this issue by starting joint training early, at epoch 100. 

This is also the reason we find it infeasible to build an entirely fair comparison between our end-to-end training and the end-to-end training proposed by CRFasRNN \cite{zheng2015conditional}. Starting end-to-end training before the unary is fully converged is impractical with CRFasRNN, a full epoch in their model takes then \num{5} hours. Additionally the CRFasRNN implementation only supports training with a batch-size of one. We find to get best results when strictly following the training protocol proposed in \cite{zheng2015conditional}, which is using a batch size of one and a very low but constant learning rate of \num{e-13}. This makes the quantitative comparison not entirely fair, however we can conclude that the training speed of our approach offers much more flexibility, ultimately resulting in better models.

\Cref{fig:learning_end2end} shows training and validation scores of our training over time. Observe the jump in training performance as soon as end-to-end training starts at epoch 100.
\section{Conclusion}

In this work we proposed Convolutional CRFs, a novel CRF design. Adding the strong and valid assumption of conditional independence enables us to remove the permutohedral lattice approximation. This allows us to implement the message passing highly efficiently on GPUs as convolution operations. This increases training and inference speed of our CRF by two orders of magnitude. In addition we observe a modest accuracy improvement when computing the message passing exactly. Our method also enables us to easily train the Gaussian features of the CRF using backpropagation.

In future work we will investigate the potential of learning Gaussian features further. We are also going to examine more sophisticated CRF architectures, towards the goal of capturing global context information even better. Lastly we are particularly interested in exploring the potential of ConvCRFs in other structured applications such as instance segmentation and landmark recognition.

\bibliography{convcrf}
\end{document}